%% file: iris_rss_paper.tex
\documentclass[conference]{IEEEtran}
\usepackage{times}

\usepackage[numbers]{natbib}
\usepackage{multicol}
\usepackage[bookmarks=true]{hyperref}

\usepackage{amsfonts}
\usepackage{amsmath}
\usepackage{amssymb}
\usepackage{graphicx}
\usepackage{caption}
\usepackage{subcaption}
\usepackage{color}

\usepackage{algorithm}
\usepackage[noend]{algpseudocode}

\newcommand{\R}{\mathbb{R}}

\newcommand{\Fk}{\text{Fk}}
\newcommand{\map}[1]{}

\begin{document}

\title{Growing Convex Collision-Free Regions in Configuration Space using Nonlinear Programming}

\author{
	\authorblockN{Mark Petersen}
	\authorblockA{School of Engineering and Applied Sciences\\
	Harvard University, Cambridge, MA 02138--2933\\
	Email: \texttt{markpetersen@g.harvard.edu}}
	\and
	\authorblockN{Russ Tedrake}
	\authorblockA{Computer Science \& Artificial Intelligence Laboratory\\
	MIT, Cambridge, MA 02139--4309\\
	Email: \texttt{russt@mit.edu}}
}

\maketitle

\begin{abstract}
One of the most difficult parts of motion planning in configuration space is ensuring a trajectory does not collide with task-space obstacles in the environment.
Generating regions that are convex and collision free in configuration space can separate the computational burden of collision checking from motion planning.
To that end, we propose an extension to IRIS (Iterative Regional Inflation by Semidefinite programming) \cite{deits2015computing} that allows it to operate in configuration space.
Our algorithm, IRIS-NP (Iterative Regional Inflation by Semidefinite \& Nonlinear Programming), uses nonlinear optimization to add the separating hyperplanes, enabling support for more general nonlinear constraints.
Developed in parallel to \citet{amice2023finding}, IRIS-NP trades rigourous certification that regions are collision free for probabilistic certification and the benefit of faster region generation in the configuration-space coordinates.
IRIS-NP also provides a solid initialization to C-IRIS to reduce the number of iterations required for certification.
We demonstrate that IRIS-NP can scale to a dual-arm manipulator and can handle additional nonlinear constraints using the same machinery.
Finally, we show ablations of elements of our implementation to demonstrate their importance.
\end{abstract}

\IEEEpeerreviewmaketitle

\section{Introduction \& Related Work}
\label{sec:intro}
\input{sections/introduction}

\section{Technical Approach}
\label{sec:approach}
\input{sections/approach}

\section{Implementation}
\label{sec:implementation}
\input{sections/implementation}

\section{Experiments}
\label{sec:experiments}
\input{sections/experiments}

\section{Conclusion} 
\label{sec:conclusion}
\input{sections/conclusion}



\bibliographystyle{plainnat}
\bibliography{references}

\end{document}

%% file: sections/introduction.tex
One of the fundamental challenges of collision-free motion planning is ensuring the entire trajectory remains safely outside the obstacles.
While checking that an individual point is collision free can be done efficiently, verifying that an entire trajectory is outside of every obstacle is a much harder problem.
In addition, decomposing the free space into overlapping convex sets has provided one solution for formulating the planning problem in a way that can be solved to global optimality \cite{deits2015efficient}.
However, performing that decomposition in configuration space is non-trivial as obstacles that are convex in task space can become nonconvex when mapped into the configuration space of the robot.

This paper extends the original IRIS (Iterative Regional Inflation by Semidefinite programming) algorithm proposed by \citet{deits2015computing} to compute convex, collision-free regions in configuration space.
IRIS relies on the assumption that the obstacles are convex.
This works well when looking for task-space collision-free regions in the presence of task-space convex obstacles.
But when the user needs convex regions in configuration space and the description of the obstacles are only convex in task space, IRIS does not work.
The algorithm does not consider the nonlinear kinematics mapping between task and configuration space, requiring task-space obstacles to be explicitly mapped into configuration space which thus far has proven to be an intractable problem.

\begin{figure}[t]
\centering
\includegraphics[width=0.35\textwidth]{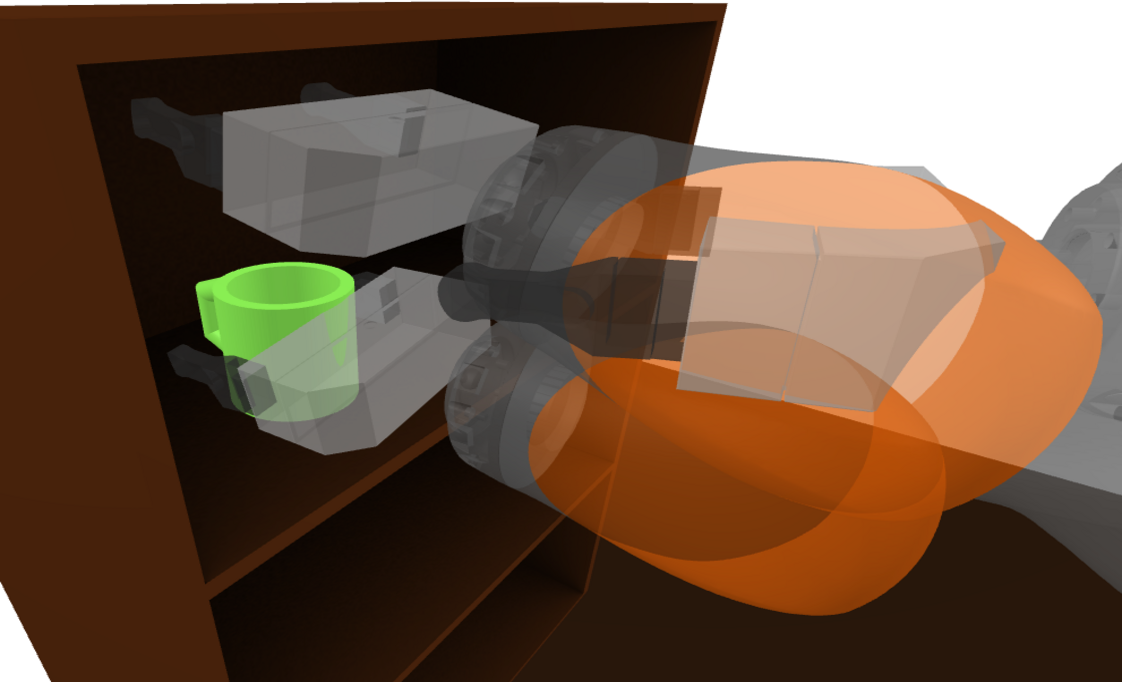}
\caption{Three configuration that lie in the same convex collision-free region of configuration space generated by IRIS-NP.  Convex regions that contain both approaching grasp and grasping configurations allow users to plan all the way up to grasp efficiently, without requiring the heuristic pre-grasp configuration used by other planners.
}
\label{fig:refined_region}
\end{figure}

Instead of explicitly mapping task-space obstacles to configuration space, we present a method to utilize an implicit configuration-space representation of the obstacles through forward kinematics.  Our algorithm, IRIS-NP (Iterative Regional Inflation by Semidefinite \& Nonlinear Programming) contributes a generalization of the iterative process of IRIS by replacing the convex problem for adding hyperplanes with a nonlinear problem. By doing this we not only can handle avoiding task-space obstacles while moving in configuration space, we are also able to handle additional nonlinear constraints on the configuration of the robot.

An alternate extension to IRIS that considers the robot kinematics for converting between configuration space and task space is C-IRIS (C-space Iterative Regional Inflation by Semidefinite programming), laid out in \citet{amice2023finding}.
C-IRIS utilizes a reparameterization of the robot configuration to convert the forward-kinematics to a rational function that can be optimized using  Sums-of-Squares (SOS) programming.
By formulating each step as a convex optimization that can be solved to global optimality, C-IRIS is able to provide rigorous guarantees that the entire region is collision free.
Since IRIS-NP uses nonlinear optimization to handle the forward-kinematics the region is convex in the configuration-space coordinates but can only be probabilistically certified collision free.
Empirically though, we find very few colliding configurations within these regions and the probability of colliding configurations can be reduced by solving the nonlinear optimization from multiple initial guesses.
Also, IRIS-NP can be used to seed C-IRIS by generating an initial region in the reparameterized coordinates.
C-IRIS can then adjust the region to quickly find certified collision-free regions.

At a high level, IRIS-NP starts with an initial seed that is a point in $\R^n$ and a polytope defined only by the joint limits. On each iteration, we formulate a nonlinear optimization that searches for configurations within the current polytope that are in collision, an optimization that we refer to as counterexample search.
These conterexamples are used to define the polytope of collision-free configuration space (C-free).
Once all obstacles have been checked, a volume-maximizing ellipse is fit to the inside of the polytope to defines a metric which maximizes volume for the next iteration of counterexample searches.

The problem of decomposing a non-convex space into a collection of convex regions has attracted several different approaches.
Many of these seek to exactly decompose a non-convex shape into approximately convex components.
\citet{lien2004approximate} performs the decomposition by iteratively splitting a non-convex shape to remove the largest concavity. 
\citet{liu2010convex} instead formulates the problem as a mixed-integer optimization to find the best cuts to break the shape into components with concavity below a given threshold.
\citet{mamou2009simple} clusters the faces of the shape to find faces that together form components of the decomposition that are approximately convex.
All of these methods return only approximately convex components that cover the original shape.
Taking the convex hull of these shapes and using that for motion planning could result in regions that intersect with obstacles.
We instead search for an inner approximation of C-Free in this work.
In addition, all of these methods require a mesh representation of the space to decompose.
In the case of generating configuration-space regions among task-space obstacles, this requires not only mapping the obstacles from task to configuration space, which is intractable to do explicitly, but also finding the complement of the resulting configuration-space obstacle mesh.

The ability to generate convex, collision-free regions in configuration space using IRIS-NP is a useful tool for motion planning and can be used directly in algorithm such as \citet{marcucci2022motion}.
These regions could also be used for collision checking in sampling-based motion planners, eliminating the need to check for collisions at regular intervals along each segment, instead directly checking that the entire segment is safe.

The remainder of this paper digs into how IRIS-NP is implemented and performs. In Section \ref{sec:approach}, we discuss the mathematical formulation for our algorithm and the iterative sequence of optimizations that are solved to grow a region.
We continue in Section \ref{sec:implementation} by discussing details of our implementation that improve runtime, simplify the resulting polytope, and extend the capabilities of IRIS-NP.s
In Section \ref{sec:experiments}, we demonstrate how well IRIS-NP region generation works and compare how the implementation details affect performance.

\textbf{Notation:}
In this paper, we will use capital letters ($X$) to denote matrices and lower case letters ($x$) to denote vectors. The monogram notation used in \citet{manipulation}, which attaches relative-to and expressed-in frames to spatial vectors, will be used for rigid transforms.

%% file: sections/approach.tex
IRIS-NP mirrors IRIS by searching for the convex polytope with largest volume inscribed ellipse.
While the true goal is to generate a polytope of maximum volume, calculating the volume of an polytope is NP-hard.
Calculating the volume of an ellipse can be done with convex optimization and maximizing the volume of the largest inscribed ellipse achieves a similar goal as maximizing the polytope's volume.
As in \citet{deits2015efficient}, we represent the ellipse as the image of unit ball
$ \mathcal{E}(C,d) =  \{x | (x-d)^T C^T C (x-d) \leq 1 \}$
and the polytope as a collection of halfplanes
$P(A, b) = \{x | Ax \leq b\}$.
Put together the optimization we would like to solve is
\begin{subequations}
\begin{align}
\label{eq:full_cost}
\min_{A, b, C, d} \quad & \text{det} \ C \\
\label{eq:full_fk}
\begin{split}
\text{s.t.} \quad & \Fk_{\mathcal{O}_i} (q,\: ^{O_i}p^x) \neq \Fk_{\mathcal{O}_j } (q,\: ^{O_j}p^y) \\
	& \quad \quad \forall \ ^{O_i}p^x \in \mathcal{O}_i,\ ^{O_j}p^y \in \mathcal{O}_j,\ q \ | \ Aq \leq b
\end{split}\\
\label{eq:full_poly}
& Ax \leq b \quad \forall \ x \ | \ (x-d)^T C^T C (x-d) \leq 1
\end{align}
\end{subequations}

\noindent
where $\mathcal{O}_i$ and $\mathcal{O}_j$ represent a pair of collision geometries and is applied for all collision pairs $i, j$ in the set of possible collision pairs $\mathcal{C}$.
The set of possible collision pairs includes both collisions between the robot and the world as well as collision between the robot and itself.
The notation $ ^{O_i}p^x$ describes a 3 dimensional vector corresponding to the position of $x$ relative to the origin of body $O_i$, expressed in the frame of $O_i$.
$\Fk_{\mathcal{O}_i} (q,\: ^{O_i}p^x)$ maps the point $x$ from the representation relative to $O_i$ to a representation relative to the world frame when the configuration of the robot is $q$.

This optimization will maximize the total volume of the ellipse \ref{eq:full_cost}.
The first constraint \ref{eq:full_fk} ensures that, for each pair of collision bodies, there is no configuration in the polytope $P$ where for any two bodies, $\mathcal{O}_i$ and $\mathcal{O}_j$, there exists points $ ^{O_i}p^x \in \mathcal{O}_i,\ ^{O_j}p^y \in \mathcal{O}_j$ that are coincident.
The second constraint \ref{eq:full_poly} ensures that the ellipse is fully contained within the polytope.
Taken together, these ensure that the polytope is as large as possible while still separating its interior from the configurations that lead to collision.
Since the forward kinematics for most robots are nonlinear, this optimization is even more difficult to solve than the one given by Equation 1 in \citet{deits2015computing}.
Specifically, the nonlinear portion of this problem can only be solved with local optimization instead of the convex optimization that was used for the convex problems that made up the iterative steps of IRIS.
Despite this, we can still use the same iterative process for splitting up the polytope optimization and the ellipse optimization.
While only using local nonlinear optimization prevents us from making guarantees that we have found a globally-optimal solution, in practice we find that we can still find high-quality solutions that achieve the goal of maximizing the volume of the collision-free polytope.

\begin{algorithm}[!h]
\caption{Given a seed point $q_0$, a list of collision pairs $\mathcal{C}$, and a bounding box on the regions limits (usually the robots joint limits) $q_{upper}$ \& $q_{lower}$ for the upper and lower limits respectively. Find a polytope $P = \{x | Ax \leq b\}$ and ellipse $\mathcal{E}(C,d) =  \{x | (x-d)^T C^T C (x-d) \leq 1 \}$ such that $\mathcal{E} \subset P$ and no collision pair in $\mathcal{C}$ are in collision for any configuration in $P$. The $AddSeparatingHyperplanes$ method is expanded in Algorithm \ref{alg:hyperplane}. The $InscribedEllipse$ method is explained in Section 3.4 of \citet{deits2015computing}.}\label{alg:iris_nl}
\begin{algorithmic}
\State $A_0, \ b_0 = [I, -I]^T, \ [q_{upper}, -q_{lower}]^T$
\State $C_0, \ d_0 = \epsilon I, \ q_0$
\State $i = 0$
\While{$i < \text{iteration limit}$}
	\State $(A_{i+1}, b_{i+1}) = AddSeparatingHyperplanes(\mathcal{E}_i, \mathcal{C}, P_0)$
	\State $(C_{i+1}, d_{i+1}) = InscribedEllipse(A_{i+1}, b_{i+1})$
	\State $i = i + 1$
	\If{($\text{det} C_i - \text{det} C_{i-1}) / \text{det} C_{i-1} < tolerance$}
    	\State \textbf{break}
	\EndIf
\EndWhile
\Return $A_i,\ b_i$
\end{algorithmic}
\end{algorithm}

\subsection{Initializing the Algorithm}
As in \cite{deits2015computing}, the algorithm starts with an initial seed configuration, $q_0$, that is not in collision around which the region will grow.
Using the seed, the polytope $P_0$ can be initialized using the joint limits of the robot and the ellipse $\mathcal{E}_0$ can be initialized as a hypersphere with small radius $\epsilon$ centered about the initial seed.

\subsection{Adding Separating Hyperplanes}

\begin{figure}[t]
\centering
\includegraphics[width=0.4\textwidth]{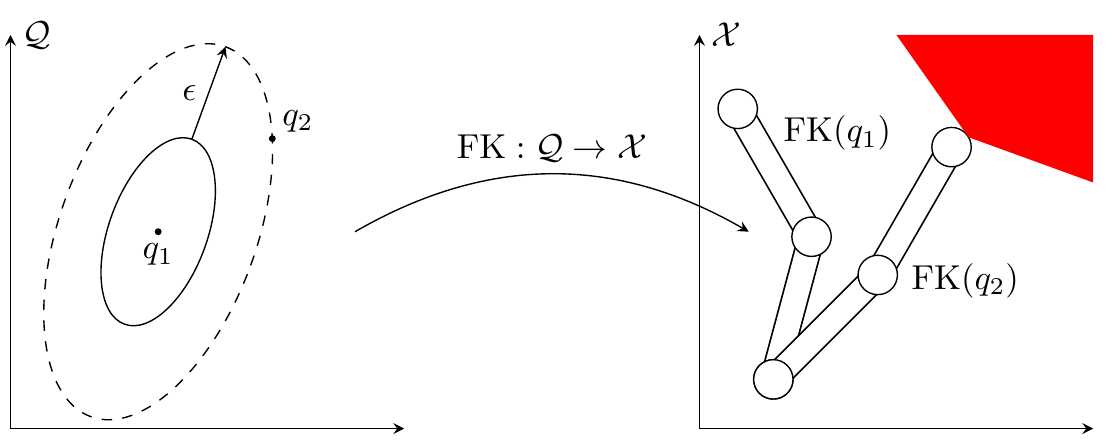}
\caption{The counterexample search consists of finding the first configuration on a uniform expansion of the ellipse that results in collision.
}
\label{fig:counterexample}
\end{figure}

In order to determine what hyperplanes to add to the polytope, the algorithm iterates over all pairs of collision bodies and searches for configurations within the polytope that result in collision.
By finding these counterexamples, tangent planes can be added to the polytope $P$ to separate the collision-free configurations from the in-collision configurations.
Conceptually, the process of finding these counterexamples consists of uniformly expanding the ellipse until a configuration on the surface of the ellipse is in collision.
A visualization of this idea is shown in Figure \ref{fig:counterexample}.

For this paper, we assume the environment is known with all collision geometries of both the robot and obstacles defined as convex sets in task space;
all non-convex geometries have been decomposed into convex components.
While IRIS-NP does not require that all collision geometries are convex in task space, having only convex collision geometries makes two of the constraints in the counterexample search convex, reducing the difficulty of the problem.

The counterexample search can be written down as

\begin{subequations}
\label{eq:counterexample}
\begin{align}
\label{eq:hyp_cost}
\min_{q, ^{O_i}p^x, ^{O_j}p^y} \quad & (q-d)^T C^T C (q-d) \\
\label{eq:hyp_points}
\text{s.t.} \quad & ^{O_i}p^x \in \mathcal{O}_i , \quad ^{O_j}p^y \in \mathcal{O}_j \\
\label{eq:hyp_fk}
& ^WX^{O_i}(q) \: \cdot \, ^{O_i}p^x = \, ^WX^{O_j}(q) \: \cdot \, ^{O_j}p^y \\
\label{eq:hyp_poly}
& A q \leq b
\end{align}
\end{subequations}

\noindent
The cost for this optimization \ref{eq:hyp_cost} specifies the configuration that is closest to the center of the ellipse using the distance metric, $C$, given by the ellipse $\mathcal{E}$. The convex constraints \ref{eq:hyp_points} select points, $^{O_i}p^x$ and $^{O_j}p^y$, that respectively lie within the geometries $\mathcal{O}_i$ and $\mathcal{O}_j$ of the collision pair. \ref{eq:hyp_poly} ensures that only configurations within the current polytope are considered. Lastly, the constraint \ref{eq:hyp_fk} specifies that the configuration $q$, when passed through the forward kinematics, results in points $x$ and $y$ being coincident in the world frame.
All of the costs and constraints for this problem are convex, except for the kinematic constraint \ref{eq:hyp_fk}. The forward kinematics make this a nonlinear optimization which can be solved using an off-the-shelf nonlinear solver.
This problem returns a feasible solution only when the polytope contains a configuration that results in collision.

\begin{figure}[t]
\centering
\begin{subfigure}[t]{0.23\textwidth}
\centering
\includegraphics[width=\textwidth]{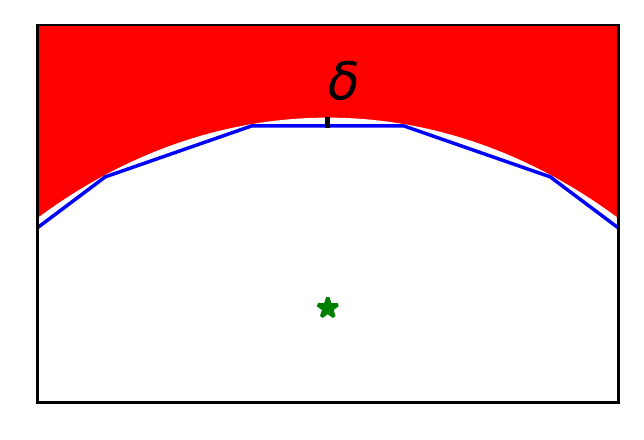}
\end{subfigure}
\
\begin{subfigure}[t]{0.23\textwidth}
\centering
\includegraphics[width=\textwidth]{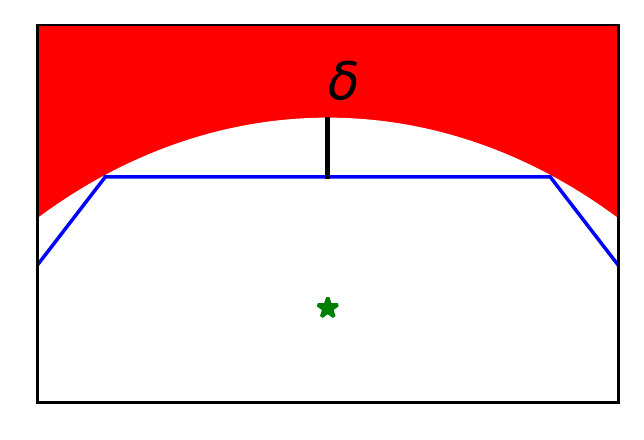}
\end{subfigure}
\caption{To separate obstacles that are concave in configuration space from the interior of the polytope with finitely many hyperplanes, the configuration-space margin backs the hyperplane away from the surface of the obstacle. Increasing this margin reduces the number of faces in the final polytope at the expense of a more conservative region.
}
\label{fig:config_margin}
\end{figure}

Once a counterexample has been found, we add a hyperplane to the iteration's polytope $P_i$ to ensure that other configurations that are in collision are excluded from the polytope.
If the obstacle is convex in configuration space, a hyperplane that is tangent to an expansion of the ellipse at the counterexample point would fully separate the interior of the polytope from collision.
As in \citet{deits2015computing} such a hyperplane can be defined using the counterexample point and the gradient of the ellipse's boundary at that point

\begin{equation}
\label{eq:hyperplane_no_margin}
a_j = C^T C (q-d), \quad b_j = a_j q.
\end{equation}

However, we cannot assume that the obstacle is convex in the configuration space.
As a result, we employ a configuration-space margin, $\delta$, that backs the hyperplane away from the obstacle by a user defined margin.
In our implementation, $\delta$ takes the units of distance in the configuration space, moving the hyperplane away from the obstacle by a fixed amount.
This makes the hyperplane more conservative than necessary but ensures that a finite number of hyperplanes can separate the obstacle from the interior of the polytope as shown in Figure \ref{fig:config_margin}.  We use the same formula as Equation \ref{eq:hyperplane_no_margin} but normalize the normal vector $a_j$ and subtract the configuration-space margin from $b_j$

\begin{equation}
\label{eq:hyperplane}
a_j = \frac{C^T C (q-d)}{||C^T C (q-d)||}, \quad b_j = a_j q - \delta.
\end{equation}
The counterexample search is performed repeatedly with the updated polytope and the same collision pairs until the program returns infeasible, ensuring that enough hyperplanes have been added to separate the collision geometries.
We repeat this search for each pair of collision geometries.
However, with the counterexample search being a nonlinear optimization, the solver reporting infeasibility does not guarantee that it is globally infeasible, meaning that a colliding configuration could escape the counterexample search. A method for reducing the probability of this happening is discussed in Section \ref{sec:prob_cert}.

\begin{algorithm}[!h]
\caption{\textbf{AddSeparatingHyperplanes} Given an ellipse defined by $C$, $d$, add separating hyperplanes to the polytope $P_i(A,b)$ that are tangent to the expanded ellipse to prevent collision pairs in the sorted list $\mathcal{C}$ from colliding.}\label{alg:hyperplane}
\begin{algorithmic}
\For {$\mathcal{O}_i$, $\mathcal{O}_j$ in $\mathcal{C}$}
	\State Setup counterexample optimization Equation \ref{eq:counterexample}
	\State $failures = 0$
	\While{$failures < \text{max infeasible samples}$}
		\If {Solve counterexample successful $\rightarrow q*$}
			\State $a_j = \frac{C^T C (q-d)}{||C^T C (q-d)||}$
			\State $b_j = a_j q - \delta$
			\State Add $a_j$, $b_j$ to $A$, $b$
			\State $failures = 0$
		\Else
			\State $failures = failures + 1$
		\EndIf
	\EndWhile
\EndFor
\Return $A,\ b$
\end{algorithmic}
\end{algorithm}

\subsection{Calculating the Largest Inscribed Ellipse}
Once the counterexample search has been completed for each collision pair, the result is a convex collision-free polytope in configuration space.
The next step of the iterative process is to find the inscribed ellipsoid of maximum volume.
Given the representation of the ellipse $ \mathcal{E}(C,d) =  \{x | (x-d)^T C^T C (x-d) \leq 1 \}$
and the polytope $P(A, b) = \{x | Ax \leq b\}$, if we define $C = \tilde{C}^{-1}$ the optimization we want to solve is

\begin{equation}
\begin{aligned}
\max_{\tilde{C}, d} \quad & \log \text{det} \tilde{C} \\
\text{s.t.} \quad & ||a_i\tilde{C}||_2 \leq b_i - a_id, \ \forall i \\
& \tilde{C} \succeq 0.
\end{aligned}
\end{equation}

\noindent
as stated by \citet{boyd2004convex} where $a_i$ are the rows of $A$ and $b_i$ are the elements of $b$.  This is a convex optimization (a semidefinite program to be precise) that can be efficiently solved using commercial solvers. For our implementation, we used both Mosek \cite{mosek} and Gurobi \cite{Gurobi} and found comparable performance from each of them.

\subsection{Termination criteria}
As with the original IRIS algorithm, this algorithm will converge to a maximal size as the volume of the inscribed ellipsoid is monotonically increasing and bounded by the initial joint limits.
To account for the fact that no bounds on the number of iterations required to achieve convergence are currently known, multiple termination criteria are provided for the algorithm.
These include:
\begin{itemize}
  \item A threshold on growth rate of the inscribed ellipse's volume
  \item An iteration limit
  \item Containment of the initial seed
\end{itemize}
Taken together these ensure the algorithm terminates in a timely manner while still generating large regions.

%% file: sections/implementation.tex
While the above approach is sufficient to generate convex collision-free regions in configuration space, we found several implementation details that accelerated the region generation.
The new formulation also made it possible to support additional constraints while generating regions.
In this section we first dive into implementation details that affect the runtime and accuracy of the region. Then we look at how this algorithm can support novel constraints.
A code implementation that supports all of these features will be made publicly available for the final version.

\subsection{Ordering Collision Pairs}
\label{sec:ordering}
As in the original IRIS algorithm, the order in which collision pairs are considered has a significant impact on the runtime and number of hyperplanes added to the polytope.
Adding hyperplanes to separate close obstacles can also separate more distant obstacles, eliminating the need to add a hyperplane for that obstacle later.
If the more distant obstacle is considered first, a hyperplane will be added to separate it, and later another hyperplane will be added to separate the closer obstacle, making the first hyperplane redundant.

In the original IRIS paper, obstacles were sorted by distance from the seed point to ensure closer obstacles are considered first.
What we would like to do is sort the collision pairs in a similar manner, by the distance in configuration space from seed to collision.
However, calculating this distance is non-trivial and effectively requires solving the optimization for finding counterexamples.
As a result, we instead sort the collision pairs based on the task-space distance between the two collision bodies when the robot is in the seed configuration.
Empirically, this serves as a good heuristic for sorting the collision pairs and leads to closer collision pairs being considered first.

\subsection{Probabilistic Certification}
\label{sec:prob_cert}
IRIS-NP ensures that the convex region is collision free by searching the current polytope for configurations that cause collision pairs to intersect.
This search is done by solving a nonlinear optimization, which means in some cases, the solver may report infeasibility when a solution does exist.
One way to get around this issue is to solve the optimization with different initial guesses.
Instead of stopping the counterexample search after the first failure to solve, the search continues from different initial conditions until a user defined number of consecutive optimizations fail to find a solution.
Initial conditions are sampled from the current convex polytope using a Markov chain Monte Carlo strategy describe in \citet{belisle1998convergence} because uniform sampling within an arbitrary polytope is not feasible.
As the number of consecutive infeasible samples is increased, the probability of missing the local region about a configuration that yields collision diminishes, providing a probabilistic certification of the region.
In addition, this gives the user a knob to trade off runtime of the algorithm with strength of the collision-free guarantee.
If we assume that the nonlinear optimization has some non-empty region of attraction for ever optimal counterexample, then the Markov chain Monte Carlo sampler is sufficient to guarantee probabilistic completeness of our counterexample search.

\subsection{Support for Additional Constraints}
\label{sec:add_constraints}
Since the process of adding hyperplanes to avoid collision consists of searching for points that violate a nonlinear inequality constraint, this opens the door for IRIS-NP to support general nonlinear inequality constraints.
Any constraint of the form $g(q) \leq 0$ is supported and counterexamples that violate the constraint are searched for using the optimization

\begin{equation}
\begin{aligned}
\min_{q} \quad & (q-d)^T C^T C (q-d) \\
\text{s.t.} \quad & g(q) \geq 0 \\
& A q \leq b.
\end{aligned}
\end{equation}

\noindent
This formulation can be used to support constraints on such things as the orientation of a kinematic frame, the position of an end-effector, or the distance between two task-space points.
So long as each is written as an inequality constraint and the feasible set has a sufficiently large interior, the mechanics presented here work well.
Equality constraints cannot be supported as they collapses the region to lie on the lower dimensional surface of the constraint, making the region zero volume.
Just as with the collision avoidance constraints, solving repeatedly from different initial conditions until a set number of consecutive optimizations have failed decreases the chances of missing a counterexample, helping to ensure the constraint is satisfied everywhere within the polytope.

\subsection{Achieving Coverage with Multiple Regions}
\label{sec:cspace_obstacles}
Up to this point, we have only considered generating individual collision-free regions.
Generating multiple regions that provide an approximate cover of C-free can be useful for downstream planning problems \cite{marcucci2022motion}.
When growing individual regions, the explicit goal is to grow them as large as possible.
While this is desirable for a singular region, this can result in regions grown from different seeds expanding to fill the same open space, away from tighter crevices that are near to each seed.
While each region is optimizing for coverage, the result is that the total coverage of all the regions is not much larger than any of the individual regions.
The goal that we'd like to achieve is maximizing the coverage of all the regions collectively.

The heuristic we use to achieve this goal is to explicitly reduce the overlap between regions.
After generating one region, that region is treated as a configuration-space obstacle.
We can then interleave the process for adding separating hyperplane from \citet{deits2015computing} with the process for adding separating hyperplanes for obstacles discussed above.
As a result, the next polytope is prevented from overlapping with the previously found regions while remaining collision free.
Adding this to our iterative algorithm encourages the new regions to expand into spaces that have not already been covered by existing regions, increasing total coverage.
Other heuristics for achieving this goal may perform better as overlapping regions can have better coverage with fewer regions\footnote{Imagine a collision-free space in the shape of a cross. Using overlapping convex regions, the entire space can be covered with two regions. If the convex regions are required not to overlap, three regions are needed to achieve complete coverage.}.
Searching for better heuristics is an active area of study.

\subsection{Region Refinement for Runtime Obstacles}
\label{sec:region_refinement}
While IRIS-NP can efficiently calculate convex collision-free regions, as soon as collision geometry changes in a way not specified by a change in configuration, the region may no longer be collision free.
While regions could be regenerated from seed points, this may not be the most efficient approach, especially if the change in collision geometry is small.
An alternate approach is to take the original region and refine it by adding additional hyperplanes to separate the new collisions.

Using the final ellipsoid as the starting metric, a single iteration of adding hyperplanes can be performed, keeping the hyperplanes that defined the original region.
Using the final ellipsoid as the starting metric ensures the overall shape of the region is leveraged during the refinement.
Alternatively, if refining the region using the final ellipse results in a configuration of interest (such as the start or goal state for a planning problem) no longer being contained in the refined region, that configuration can be used as an additional seed point to grow a new region inside the original region.
Only collision pairs involving the changed geometries need to be considered as the remaining collision pairs were already separated during the original region generation.
This approach is well suited to the case when novel geometry that had not been previously considered is added to the scene.
If geometry is removed from the scene or moved to be less restrictive of the robot's motion, the new parts of C-free will not be covered by existing regions and the refinement process will not cover that gap.s

This refinement approach was used to generate the region shown in Figure \ref{fig:refined_region}.
An initial region was generated inside the shelf without any mugs in the scene.
Then a mug was added to the scene and the region was refined using a seed point with the gripper around the mug prepared to grasp.
A single iteration with the initial hypersphere about the gripping position seed point as the metric for adding hyperplanes was able to cut down the region and remove the portions that were in collision.
As is shown in Figure \ref{fig:refined_region}, refining the region to be more restrictive still left the region large enough to contain both grasping and pre-grasp configurations.
We found this surprising!
Using IRIS-NP a planner can plan straight to a grasp configuration without having to specify a pre-grasp waypoint to plan to first.

%% file: sections/experiments.tex
To test our implementation, we set up several experiments that demonstrate how each component performs and how IRIS-NP scales.  In all experiments shown SNOPT \cite{gill2005snopt} was used to solve the nonlinear counterexample search.

\subsection{Collision Pair Ordering Ablation}
\label{sec:ex/ordering}
To study the importance of the order in which collision pairs are considered, we compare generating regions when the collision pairs are ordered by task-space distance as described in Section \ref{sec:ordering} against generating regions when collision pairs are unordered.
For this comparison, we use a KUKA LBR iiwa with seven degrees of freedom, surrounded by randomly placed pillars as depicted in Figure \ref{fig:orient_con}.

Using 100 randomly selected seed points that do not place the robot into collision with either itself or the environment, we generated regions about each of these seed point.
Region generation was done with collision pairs sorted by their distance in task space when at the seed configuration and with collision pairs unsorted.
For the termination settings, we required that the sample point stay inside the region, had an iteration limit of 5 and a minimum growth rate between iterations of 2\%.
The configuration-space margin was set to 0.01 and we used 1 infeasible sample per collision pair.

\begin{table}[t]
\caption{Ordering the collision pairs by distance results in regions that are generated faster, with fewer faces and without significant change in the volume of the inscribed ellipse.
The average number of faces of the polytope and the average run time are shown for regions generated from 100 random seeds.
Because some sampled configurations are in very open spaces and others are in close proximity to many obstacles we compare the largest volume regions.
}
\centering
\begin{tabular}{| c | c | c | c |}
 \hline
 & Mean Polytope Faces & Mean Run Time & Maximum Volume\\ 
 \hline
 Ordered   & 146 $\pm$ 29 & 41.4 $\pm$ 11.7 & 86.02 \\  
 Unordered & 181 $\pm$ 27 & 54.2 $\pm$ 8.4 & 89.35 \\
 \hline
\end{tabular}
\label{tab:ordering}
\end{table}

The results of this ablation are shown in Table \ref{tab:ordering}.
On average the regions generated using ordered collision pairs were generated faster and had less faces than the regions generated with unsorted collision pairs.
In addition, there was not a significant difference in the volume of the maximum inscribed ellipse, suggesting that the volume of the overall regions were comparable.

\subsection{Probabilistic Certification}

\begin{figure}[t]
\centering
\includegraphics[width=0.35\textwidth]{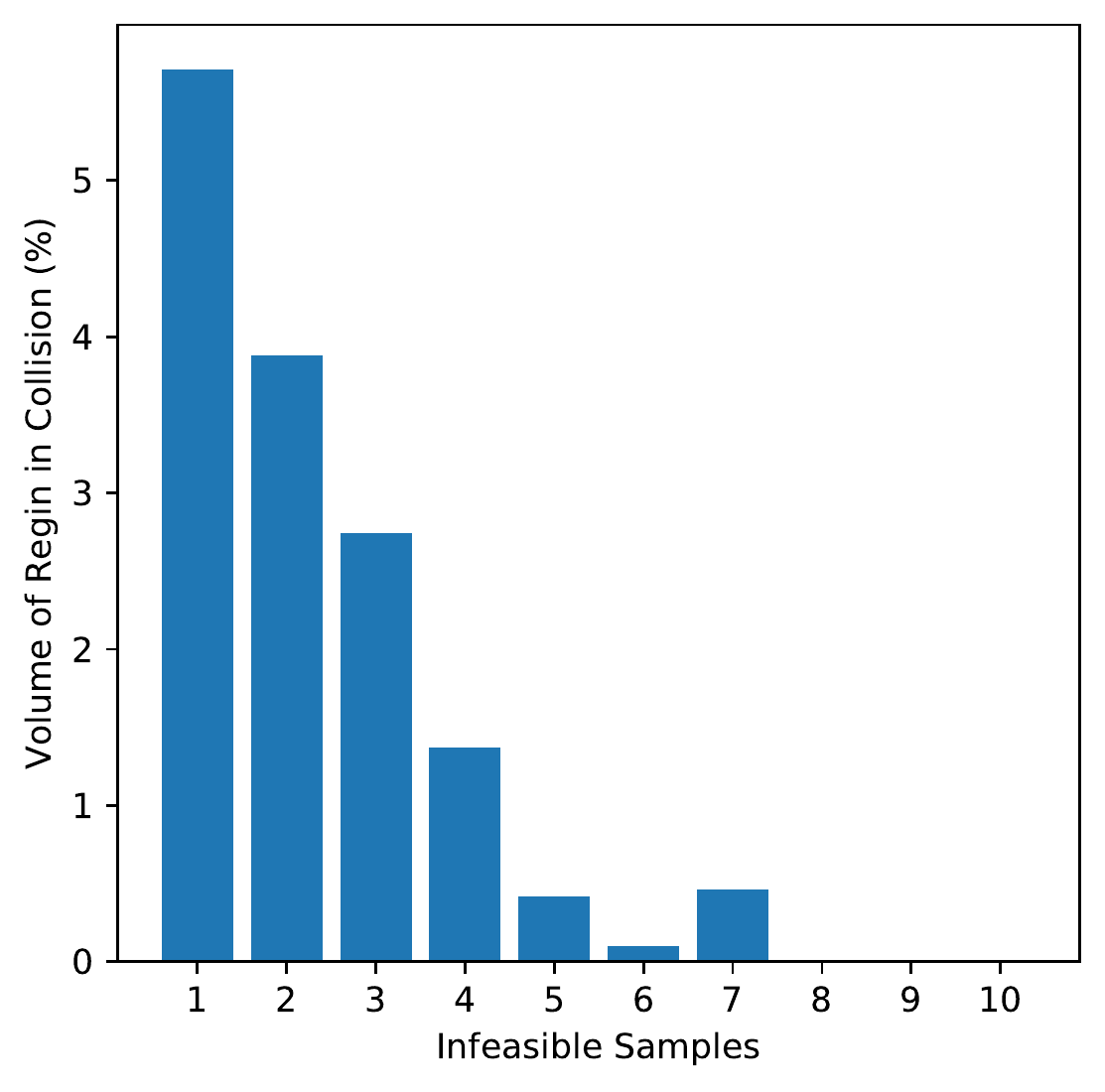}
\caption{As the number of consecutive infeasible counterexample searches required before continuing increases, the percent of samples within the region that are in collision heads to zero. This trend allows the user to trade off the run time of region generation versus probabilistic certification of the region. Note that this specific environment and seed configuration was explicitly designed to make it very difficult for IRIS-NP to generate completely collision-free regions. In most practical environments, we find that a single infeasible sample is sufficient to eliminate all collisions from the region.
}
\label{fig:certificate}
\end{figure}

For this section, we look at the ability to provide a probabilistic certificate of the regions generated with IRIS-NP.
As mentioned previously, because the process of adding hyperplanes to separate obstacles relies on a nonlinear optimization, we cannot guarantee that there are no collisions inside the region, even when the solver reports infeasible.
A possible solution is to increase the number of consecutive infeasible optimizations that are solved from randomly sampled initial guesses before ending the search for hyperplanes.

To test if this reduced the number of colliding configurations inside the region, we set up a simple environment with a 4 link arm in the plane, with circular obstacles around the robot.
The robot, environment, and seed were selected to make it very difficult to generate a region that was completely collision free.
In practice, most of the regions generated for real problems had few if any collisions in the region, even with just a single infeasible sample.
Regions were generated about the same seed configuration while varying the number of consecutive infeasible counterexample searches to perform.
For the termination settings, we required that the sample point stay inside the region, had an iteration limit of 5 and a minimum growth rate between iterations of 2\%.
The configuration-space margin was set to 0.01.
We then randomly sampled configurations within the region (using rejection sampling) to calculate the percentage of configurations within the region that are in collision.
The results are shown in Figure \ref{fig:certificate}.

As we increased the number of infeasible problems the solver must solve before moving on to a different collision pair, the percent of colliding configurations that were in the region dropped.
The the decrease in percent of colliding configurations was not monotonic but that is likely due to the interplay between adding more hyperplanes during the counterexample step and the metric used to add hyperplanes, which is changed on the major iterations of IRIS-NP when we calculate a new maximum inscribed ellipse.
We do not claim the process is monotonic, only that the percent of colliding configurations within the region converges to zero in the limit.

\subsection{Additional Constraints}

\begin{figure}[t]
\centering
\begin{subfigure}[t]{0.23\textwidth}
\centering
\includegraphics[width=\textwidth]{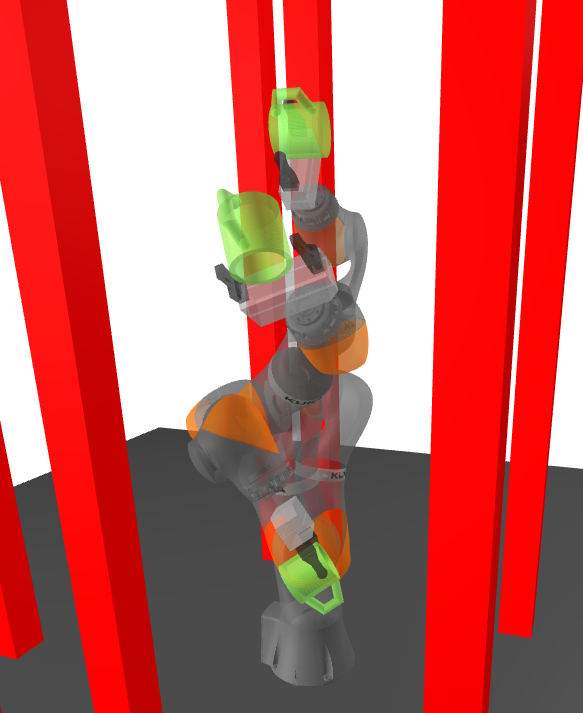}
\end{subfigure}
\
\begin{subfigure}[t]{0.23\textwidth}
\centering
\includegraphics[width=\textwidth]{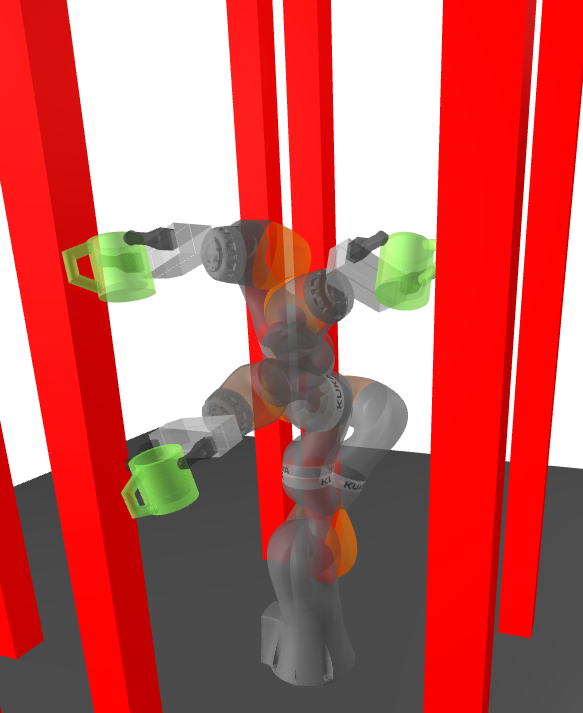}
\end{subfigure}
\caption{A comparison of the convex collision-free regions generated without (left) and with (right) an additional gripper orientation constraint designed to prevent a held mug from spilling its contents. The red pillars are task-space obstacles that both the robot and mug must not collide with.
}
\label{fig:orient_con}
\end{figure}

As mentioned in Section \ref{sec:add_constraints}, using the same machinery that is used to generate convex collision-free regions, IRIS-NP can support additional nonlinear constraints on the configuration of the robot.
One such constraint is an orientation constraint on the end effector that prevents a grasped mug from spilling its contents. We demonstrate the support for additional constraints using the same environment as was described in Section \ref{sec:ex/ordering}.
We compared generating a region that is collision free with generating a region that has the additional constraint that the gripper's orientation must be kept within 0.15 radians of level.
Both regions were generated using a seed point where the arm is reaching around and between pillars, yielding the potential for multiple collisions.
For the termination settings, we do not require the sample point stay inside the region, the iteration limit is 5 and the minimum growth rate between iterations is 2\%. The configuration-space margin was set to 0.01 and we used 3 infeasible sample per collision pair.

A few configurations from each region are shown side by side in Figure \ref{fig:orient_con}.
IRIS-NP is able to quickly optimize a region that obeys the constraint, keeping the gripper close to level and the held mug upright everywhere inside the region.
As expected, the region without the gripper orientation constraint is larger as there are more configurations that are collision-free but violate the orientation constraint.
Unexpectedly, the region with the added constraint is generated faster than the region without it, 110 seconds and 129 seconds respectively.
This is not due to the number of faces added to the polytope, as the region with the constraint has more at 576 faces than the one that does not at 172 faces.
The speedup is likely due to the fact that the added constraint quickly limited the set of configurations that could lie inside the region, requiring fewer iterations to maximize the volume of the region.

\subsection{Scaling to High Dimensions}
\begin{figure}[t]
\centering
\includegraphics[width=0.35\textwidth]{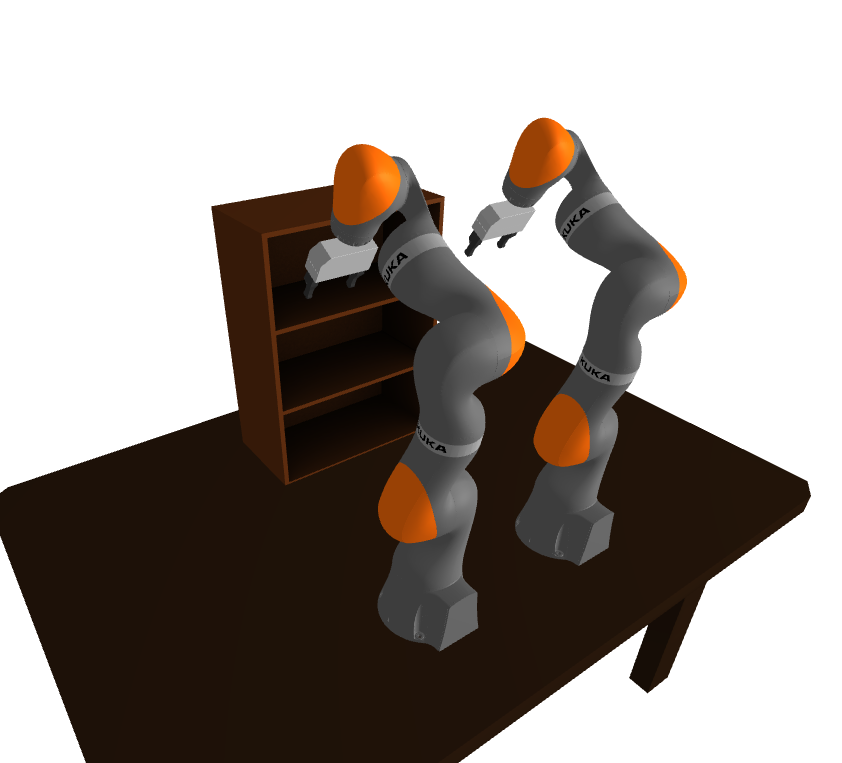}
\caption{A bimanual environment consisting of two KUKA LBR iiwa and a shelving unit. IRIS-NP can scale well to this 14 dimensional environment, constructing some regions in as little as 1 minute.
}
\label{fig:bimanual}
\end{figure}

Since IRIS-NP relies on a local nonlinear optimization to grow large convex regions, it can scale well to robots with larger numbers of degrees of freedom. To demonstrate this, we generate regions for a bimanual manipulator consisting of two KUKA LBR iiwa for a total of fourteen degrees of freedom. The environment, shown in Figure \ref{fig:bimanual} contains a shelving unit that the arms can reach into. For this environment, the region must not only confirm that neither arm is in collision with any of the environment obstacles, it must also confirm that the two arms do not collide with each other.
IRIS-NP generates large regions in this environment in 20 minutes for large open regions of the configuration space and as fast as 1 minute for the more constrained regions in the shelves.

%% file: sections/conclusion.tex
In this work, we have demonstrated an extension to IRIS to allow it to generate convex collision-free regions in configuration space when the mapping to task space is nonlinear.
These regions can be generated efficiently and the algorithm, IRIS-NP, scales to high-degree-of-freedom manipulators.
The regions generated by IRIS-NP are incredibly valuable for planning motions \cite{marcucci2022motion} and can be made probabilistically certified.
We've also shown how this process can handle additional nonlinear constraints on the configuration and how regions can be refined to handle changes to the environment.

While the ability to refine these regions does enable adapting regions to changes in the collision geometry, the current refinement process only works well for changes that add collision geometry or shrink existing regions of C-free. Handling the cases where objects move significantly or even leave the scene, without having to regenerate the regions from scratch is a focus of future work.  In addition, \citet{marcucci2022motion} demonstrated that the amount of overlap between regions created trade offs between coverage of C-free and planning time. Understanding this trade off, and how to generate regions optimally for the downstream pipeline is something we continue to explore.

%% file: iris_rss_paper.bbl
\begin{thebibliography}{13}
\providecommand{\natexlab}[1]{#1}
\providecommand{\url}[1]{\texttt{#1}}
\expandafter\ifx\csname urlstyle\endcsname\relax
  \providecommand{\doi}[1]{doi: #1}\else
  \providecommand{\doi}{doi: \begingroup \urlstyle{rm}\Url}\fi

\bibitem[Amice et~al.(2023)Amice, Dai, Werner, Zhang, and
  Tedrake]{amice2023finding}
Alexandre Amice, Hongkai Dai, Peter Werner, Annan Zhang, and Russ Tedrake.
\newblock \href{https://arxiv.org/pdf/2205.03690.pdf}{Finding and Optimizing
  Certified, Collision-Free Regions in Configuration Space for Robot
  Manipulators}.
\newblock In \emph{International Workshop on the Algorithmic Foundations of
  Robotics}, pages 328--348. Springer, 2023.
\newblock URL \url{https://arxiv.org/pdf/2205.03690.pdf}.

\bibitem[ApS(2022)]{mosek}
MOSEK ApS.
\newblock \emph{MOSEK Optimizer API for C 10.0.34}, 2022.
\newblock URL \url{https://docs.mosek.com/10.0/capi/index.html}.

\bibitem[Boyd and Vandenberghe(2004)]{boyd2004convex}
Stephen~P Boyd and Lieven Vandenberghe.
\newblock \emph{Convex optimization}.
\newblock Cambridge university press, 2004.

\bibitem[Bélisle et~al.(1998)Bélisle, Boneh, and
  Caron]{belisle1998convergence}
Claude Bélisle, Arnon Boneh, and Richard Caron.
\newblock Convergence properties of hit-and-run samplers.
\newblock \emph{Communications in Statistics. Stochastic Models}, 14, 01 1998.
\newblock \doi{10.1080/15326349808807500}.

\bibitem[Deits and Tedrake(2015{\natexlab{a}})]{deits2015computing}
Robin Deits and Russ Tedrake.
\newblock
  \href{https://link.springer.com/chapter/10.1007/978-3-319-16595-0_7}{Computing
  large convex regions of obstacle-free space through semidefinite
  programming}.
\newblock In \emph{Algorithmic foundations of robotics XI}, pages 109--124.
  Springer, 2015{\natexlab{a}}.
\newblock URL
  \url{https://link.springer.com/chapter/10.1007/978-3-319-16595-0_7}.

\bibitem[Deits and Tedrake(2015{\natexlab{b}})]{deits2015efficient}
Robin Deits and Russ Tedrake.
\newblock
  \href{https://dspace.mit.edu/bitstream/handle/1721.1/101082/Tedrake_Efficient%20mixed.pdf?sequence=1&isAllowed=y}{Efficient
  mixed-integer planning for UAVs in cluttered environments}.
\newblock In \emph{2015 IEEE international conference on robotics and
  automation (ICRA)}, pages 42--49. IEEE, 2015{\natexlab{b}}.
\newblock URL
  \url{https://dspace.mit.edu/bitstream/handle/1721.1/101082/Tedrake_Efficient%20mixed.pdf?sequence=1&isAllowed=y}.

\bibitem[Gill et~al.(2005)Gill, Murray, and Saunders]{gill2005snopt}
Philip~E Gill, Walter Murray, and Michael~A Saunders.
\newblock
  \href{https://ccom.ucsd.edu/~optimizers/static/pdfs/siam_44609.pdf}{SNOPT: An
  SQP algorithm for large-scale constrained optimization}.
\newblock \emph{SIAM review}, 47\penalty0 (1):\penalty0 99--131, 2005.

\bibitem[{Gurobi Optimization, LLC}(2023)]{Gurobi}
{Gurobi Optimization, LLC}.
\newblock {Gurobi Optimizer Reference Manual}, 2023.
\newblock URL \url{https://www.gurobi.com}.

\bibitem[Lien and Amato(2004)]{lien2004approximate}
Jyh-Ming Lien and Nancy~M Amato.
\newblock \href{https://dl.acm.org/doi/abs/10.1145/997817.997823}{Approximate
  convex decomposition of polygons}.
\newblock In \emph{Proceedings of the twentieth annual symposium on
  Computational geometry}, pages 17--26, 2004.
\newblock URL \url{https://dl.acm.org/doi/abs/10.1145/997817.997823}.

\bibitem[Liu et~al.(2010)Liu, Liu, and Latecki]{liu2010convex}
Hairong Liu, Wenyu Liu, and Longin~Jan Latecki.
\newblock \href{https://cis.temple.edu/~latecki/Papers/CSD_CVPR10.pdf}{Convex
  shape decomposition}.
\newblock In \emph{2010 IEEE Computer Society Conference on Computer Vision and
  Pattern Recognition}, pages 97--104. IEEE, 2010.
\newblock URL \url{https://cis.temple.edu/~latecki/Papers/CSD_CVPR10.pdf}.

\bibitem[Mamou and Ghorbel(2009)]{mamou2009simple}
Khaled Mamou and Faouzi Ghorbel.
\newblock A simple and efficient approach for 3d mesh approximate convex
  decomposition.
\newblock In \emph{2009 16th IEEE international conference on image processing
  (ICIP)}, pages 3501--3504. IEEE, 2009.

\bibitem[Marcucci et~al.(2022)Marcucci, Petersen, von Wrangel, and
  Tedrake]{marcucci2022motion}
Tobia Marcucci, Mark Petersen, David von Wrangel, and Russ Tedrake.
\newblock \href{https://arxiv.org/pdf/2205.04422.pdf}{Motion Planning around
  Obstacles with Convex Optimization}.
\newblock \emph{arXiv preprint arXiv:2205.04422}, 2022.
\newblock URL \url{https://arxiv.org/pdf/2205.04422.pdf}.

\bibitem[Tedrake(2022)]{manipulation}
Russ Tedrake.
\newblock \emph{Robotic Manipulation}.
\newblock 2022.
\newblock URL \url{https://manipulation.mit.edu/pick.html#monogram}.

\end{thebibliography}
